\def\year{2022}\relax
\documentclass[letterpaper]{article} % DO NOT CHANGE THIS
\usepackage{aaai22}  % DO NOT CHANGE THIS
\usepackage{times}  % DO NOT CHANGE THIS
\usepackage{helvet}  % DO NOT CHANGE THIS
\usepackage{courier}  % DO NOT CHANGE THIS
\usepackage[hyphens]{url}  % DO NOT CHANGE THIS
\usepackage{graphicx} % DO NOT CHANGE THIS
\usepackage{natbib}  % DO NOT CHANGE THIS AND DO NOT ADD ANY OPTIONS TO IT
\usepackage{caption} % DO NOT CHANGE THIS AND DO NOT ADD ANY OPTIONS TO IT
\DeclareCaptionStyle{ruled}{labelfont=normalfont,labelsep=colon,strut=off} % DO NOT CHANGE THIS
\nocopyright 
\title{Sketching as a Tool for Understanding and Accelerating\\ Self-attention for Long Sequences}
\author{
Yifan Chen$^{1}$\thanks{Equal contribution.}, 
Qi Zeng$^{1*}$,
Dilek Hakkani-Tur$^{2}$,
Di Jin$^{2}$,
Heng Ji$^{1}$,
Yun Yang$^{1}$
}
\affiliations{
$^1$University of Illinois Urbana-Champaign  
$^2$Amazon Alexa AI \\
$^1${\tt \{yifanc10, qizeng2,  hengji, yy84\}@illinois.edu} \\
$^2${\tt \{hakkanit, djinamzn\}@amazon.com}
}

\usepackage[utf8]{inputenc}

\usepackage{etoolbox} % for hyperref bug

\usepackage{paperpkg}
\usepackage{xr}
\usepackage{comment}              
\usepackage{xparse}

\usepackage{wrapfig}
\usepackage{enumitem}

\usepackage{arydshln}               % Dashed or dotted line
\NewDocumentCommand{\heng}
{mO{}}{\textcolor{red}{\textsuperscript{\textit{Heng}}\textsf{\textbf{\small[#1]}}}}
\NewDocumentCommand{\vicki}
{mO{} }{\textcolor{blue}{\textsuperscript{\textit{Vicki}}\textsf{\textbf{\small[#1]}}}}
\NewDocumentCommand{\william}{ mO{} }{\textcolor{red}{\textsuperscript{\textit{William}}\textsf{\textbf{\small[#1]}}}}

\usepackage{color}

\NewDocumentCommand{\method}{ mO{} }{#1}

\usepackage{fixltx2e}
\usepackage{dsfont}

\begin{document}

\maketitle

\begin{abstract}

% ($\m O(n^2)$ with sequence length $n$)
% alleviate this issue by attaining an $\wt{\m O}(n)$ complexity bound ($\wt {\m O}(\cdot)$ means $\m O(\cdot)$ modulo poly-log terms).

Transformer-based models are not efficient in processing long sequences due to the quadratic space and time complexity of the self-attention modules.
To address this limitation, Linformer and Informer are proposed to reduce the quadratic complexity to linear (modulo logarithmic factors) via low-dimensional projection and row selection respectively.
These two models are intrinsically connected, and to understand their connection, we introduce a theoretical framework of matrix sketching.
Based on the theoretical analysis, we propose Skeinformer to accelerate self-attention and further improve the accuracy of matrix approximation to self-attention with three carefully designed components: column sampling, adaptive row normalization and pilot sampling reutilization.
Experiments on the Long Range Arena (LRA) benchmark demonstrate that our methods outperform alternatives with a consistently smaller time/space footprint.
% \footnote{The code will be made publicly available for research purposes.} 
\footnote{Our code is released at\\ \url{https://github.com/pkuzengqi/Skeinformer}} 
\end{abstract}

% Previous methods, including Linformer and Informer, 
% to justify their success and identify their deficiencies.
% via low-dimensional projection and row selection.

% As a popular architecture, Transformer is widely used in various natural language processing tasks,
% which utilizes the softmax self-attention modules to capture the relation among tokens in a sequence.
% Despite the unparalleled performance, self-attention suffers a quadratic $\m O(n^2)$ space and time complexity for a sequence with $n$ tokens,
% which limits the ability of Transformer to address long sequences.
% Previous works to alleviate the issue include Linformer and Informer,
% which claim to attain a $\wt{\m O}(n)$ complexity bound ($\wt {\m O}(\cdot)$ means $\m O(\cdot)$ modulo poly-log terms).
% We introduces the sketching methods as a theoretical framework to provide a justification for the success of these two models.
% Via the lens of sketching, we also notice the limitation of the existing models that they do not efficiently make use of the value matrix $\mtx{V}$ in self-attention.
% We accordingly propose a new method to accelerate self-attention, which improves upon the two models regarding the accuracy of matrix approximation to self-attention.
% Numerical experiments on the benchmark, Long Range Arena (LRA), demonstrate that with the same computational cost our method has a better performance than other common efficient Transformer-like methods.

\section{Introduction}
\label{sec:intro}

Transformer~\cite{DBLP:conf/nips/VaswaniSPUJGKP17} utilizes softmax self-attention modules to capture the dependency between tokens in a sequence and has been widely used in various Natural Language Processing tasks.
The time and space complexity of the dot-product self-attention is quadratic in the input sequence length, which restricts the largest sequence length and batch size.
To adapt transformers to long sequences, documents have to be truncated, chunked using a sliding window, or processed in parallel on multiple GPUs.
These additional operations usually cause the loss of long-range dependency and introduce additional computational costs.

In this paper, we focus on efficient self-attention methods~\cite{DBLP:journals/corr/abs-2102-03902, DBLP:conf/emnlp/QiuMLYW020, DBLP:conf/nips/ZaheerGDAAOPRWY20, DBLP:journals/corr/abs-2004-05150, DBLP:conf/iclr/KitaevKL20, DBLP:journals/tacl/RoySVG21}, 
among which Linformer~\cite{DBLP:journals/corr/abs-2006-04768} and Informer~\cite{DBLP:journals/corr/abs-2012-07436} are two representative approaches to reducing the $\m O(n^2)$ self-attention to an $\wt{\m O}(n)$ operation ($\wt {\m O}(\cdot)$ means $\m O(\cdot)$ modulo poly-log terms and $n$ is the sequence length) in both space and time complexity.
Linformer forms a low-rank factorization of the original attention by decomposing it into smaller attentions,
while Informer allows each key to only attend to a certain number of queries.
% the ProbSparse self-attention in 

% These two methods can be unified and explained as sketching methods~\cite{woodruff2014sketching}, 
% which replaces the original matrix $\mtx{B}$ with its random sketch $\mtx{B} \mtx{S}$. 
% Section~\ref{sec:sketching} provides some concrete instances of commonly used distributions for constructing the random sketching matrix $\mtx{S}$.
% Here the random sketching matrix $\mtx{S}$ usually follows a certain distribution, as we will come back shortly in Section~\ref{sec:sketching}. 

To better understand self-attention, we introduce a theoretical framework, sketching~\cite{woodruff2014sketching}, to help explain the key ideas in Informer and Linformer from the perspective of matrix approximation.
Specifically, sketching methods replace the original matrix $\mtx{B}$ with its random sketch $\mtx{B} \mtx{S}$ to reduce computations.
In Section~\ref{sec:sketching} we introduce some concrete instances of commonly used distributions for constructing the random sketching matrix $\mtx{S}$.
Furthermore, taking matrix approximation as a general guideline, 
we recognize the deficiency in Informer and Linformer, 
that they either do not fully utilize the information in the value matrix $\mtx{V}$, or deviate from the original self-attention output.
This guideline also motivates us to propose \textbf{Skeinformer} through the theoretical analysis under the sketching framework.

\iffalse
\vicki{delete the following two sentences?}
This framework allows a finer analysis of approximation error, and theoretically justifies Informer, Linformer, and our method.
Section~\ref{sec:sketching} introduces some concrete instances of commonly used distributions for constructing the random sketching matrix $\mtx{S}$.
\fi

To improve the approximation accuracy in terms of the original attention output, 
Skeinformer applies sub-sampling sketching to reduce time complexity and exploits the information from the value matrix $\mtx{V}$ with \textbf{column sampling}. 
Skeinformer also incorporates an \textbf{adaptive row normalization} step, which approximates the un-selected rows by a vector with all elements $\frac1n$ and has significantly boosted the performance of Informer.
% In addition, we find that the row normalization trick in Informer, which approximates the un-selected rows by a vector with all elements $\frac1n$, significantly boosts its performance due to the rank collapse phenomenon \cite{dong2021attention}.
% This trick is also incorporated in our method, leading to our proposed model \textbf{Skeinformer}.
In addition, we introduce a simple yet effective step, \textbf{pilot sampling reutilization}, which reuses the computation of pilot sampling to improve both the approximation accuracy and training efficiency.
Our experiments on the LRA benchmark show that Skeinformer consistently uses less space and time while achieving better accuracy than most baseline methods. 
% Our methods' attention mechanism is a drop-in replacement for the standard attention.

In summary, our contributions are twofold:
\begin{itemize}
\item We introduce sketching as a theoretical framework for analyzing and developing efficient transformers.
% Specifically, by viewing the goals of Informer and Linformer as approximate matrix multiplication, we utilize the randomized sketching theory to cover these two methods.
Specifically, the randomized sketching theory covers these two methods from the perspective of approximate matrix multiplication.
This framework connects the studies on efficient transformers and randomized sketching theory, so that future development in efficient transformers and sketching can benefit each other.

\item We propose Skeinformer as a straightforward product of the sketching framework to accelerate the training and inference of transformers. 
Skeinformer consists of three components: the initial column sampling that incorporates the information from the value matrix $\mtx{V}$ into the sampling probabilities, the adaptive row normalization  that fills un-selected columns with the averaged selected columns, and the pilot sampling re-utilization.
% \item We further identify and analyze the effect of the implicit row normalization in Informer and propose to improve it with adaptive row normalization.
\end{itemize}

\section{Related work}
\label{sec:literature}

% \subsection{Fast attention}
% \subsection{Efficient transformers}

The ability to process long sequences is critical for many Natural Language Processing tasks, including Document Summarization \cite{DBLP:conf/emnlp/XiaoC19, huang2021effattn}, Question Answering \cite{ DBLP:journals/corr/abs-2009-06097}, Information Extraction \cite{DBLP:journals/corr/abs-2104-05919, DBLP:conf/acl/DuC20, DBLP:conf/acl/EbnerXCRD20}, and Machine Translation\cite{DBLP:conf/acl/Bao0TCL20}.
% \cite{DBLP:journals/corr/abs-1810-04805, DBLP:conf/acl/LewisLGGMLSZ20, DBLP:journals/corr/abs-1907-11692, DBLP:conf/acl/DaiYYCLS19, DBLP:conf/nips/BrownMRSKDNSSAA20}   % those are regular transformers
However, the quadratic computational cost of self-attention in transformer-based models limits their application in long-sequence tasks. 
Recent methods have been proposed to accelerate attention computation by selectively attending to a subset of the tokens or with low-rank matrix approximation.
We refer readers to a survey paper on efficient transformers~\citep{DBLP:journals/corr/abs-2009-06732}  for more details on attention approximation methods.

Selective attention methods limit the scope of matrix operation with sparse attention patterns or column/row sampling methods.
BlockBERT~\cite{DBLP:conf/emnlp/QiuMLYW020} introduces sparse block structures into the attention matrix.
Sparse Transformer~\citep{DBLP:journals/corr/abs-1904-10509} introduces dilated patterns.
Big Bird~\cite{DBLP:conf/nips/ZaheerGDAAOPRWY20} proposes a combination of random, window, and global attention. 
Longformer~\cite{DBLP:journals/corr/abs-2004-05150} combines local windowed attention with task-motivated global attention.
The most related work to ours is Informer~\cite{DBLP:journals/corr/abs-2012-07436}, which allows each key to only attend to the top queries under the Kullback-Leibler divergence based sparsity measurement. 

Low-rank attention matrix approximation methods are based on the assumption of low-rank structure in the full self-attention matrix.
Linformer~\cite{DBLP:journals/corr/abs-2006-04768} compresses the size of the key and value matrices by the Johnson–Lindenstrauss transform ~\citep{johnson1984extensions}.
Performer~\cite{DBLP:journals/corr/abs-2009-14794} recognizes the attention score matrix as an empirical Gaussian kernel matrix and constructs a low-rank projection for both the query and key matrices through random Fourier features~\citep{rahimi2007random}.
Nystr\"omformer~\cite{DBLP:journals/corr/abs-2102-03902} instead utilizes \nystrom method \citep{williams2001using, drineas2005nystrom} to approximate the attention score matrix.

Some other methods follow a similar principle to decompose the attention score matrix, although they are not necessarily aiming to approximate the original self-attention itself.
The representative methods include Linear Transformer~\cite{DBLP:conf/icml/KatharopoulosV020}, which claims that the exponential transform of the dot-product in the softmax operation can be replaced by the direct matrix multiplication of the projected query and key matrices,
and Reformer \citep{kitaev2019reformer}, which forces the query and key matrices to be identical and applies locality-sensitive hashing (LSH) \citep{har2012approximate} to simplify the computation of the attention score matrix.
Those methods are effective alternatives of the original self-attention, while they do not fall into the scope of matrix approximation.
We spare the discussion of those methods in this paper.

% Our work falls into the scope of sampling methods for attention approximation. 

\section{Sketching Framework for Self-Attention}
\label{sec:task}

\begin{algorithm*}
\SetAlgoLined
\small
\KwIn{query matrix $\mtx{Q}$, key matrix $\mtx{K}$, value matrix $\mtx{V}$ (all are $n$-by-$p$), and sub-sample size $d$}
\KwOut{Attention output matrix $\mtx{R}$ with the same shape as $\mtx{V}$}
 % Initialize the matrix $\mtx{R} = 0 \in \mb R^{n \times d}$\; 
 Uniformly sample $d$ indices $j_1, \cdots, j_d$ with replacement\;
 Construct the $d \times p$ matrix $\mtx{Q}_J$ as to the index set $J \defeq \{j_k\}_{k=1}^d$, whose $k$-th row is $\mtx{Q}_{(j_k)}$\;
 Compute the matrix $\mtx{B}_J = \text{softmax}\left(\mtx{Q}_J \mtx{K}^T / \sqrt{p}\right)$ \tcp*{pilot sampling}
 Based on $\mtx{B}_J$, give the estimated sub-sampling probabilities $\{\hat p_i\}^n_{i=1}$ as in Equation~(\ref{eqn:est_prob})\;
 With $\{\hat p_i\}^n_{i=1}$ sample $d$ indices $j'_1, \cdots, j'_d$ without replacement\;
 Construct the $d$-by-$p$ matrix $\mtx{K}_{J'}$ (resp., $\mtx{V}_{J'}$) according to the indices list $J' \defeq \{j'_k\}_{k=1}^d$, whose $k$-th row is $\mtx{K}_{(j'_k)}$ (resp., $\mtx{V}_{(j'_k)}$)\;
 Compute the two matrices $\mtx{A}^{J'} = \exp\left(\mtx{Q} \mtx{K}_{J'}^T / \sqrt{p}\right)$, and $\mtx{R}_{J'} = \mtx{A}^{J'} \mtx{V}_{J'}$ \tcp*{column sampling}
 Construct a length $n$ column vector $\mtx{g}$ whose $i$-th element is $(\prod_{k=1}^d a_{i j'_k})^{\frac1d}, \forall i \in [n]$\;
 Compute the row sum vector $\mtx{d} \defeq \mtx{A}^{J'} \mtx{1}_d + (n-d) \mtx{g}$ \tcp*{adaptive row normalization}
 Denote the un-selected part of $\mtx{V}$ as $\mtx{V}_{(J')^C}$, and compute the vector $\mtx{v} = \mtx{V}_{(J')^C}^T \mtx{1}_{n-d}$\;
 Obtain the intermediate output $\mtx{R} = \text{diag}(\mtx{d}^{-1}) (\mtx{R}_{J'} + \mtx{g} \mtx{v}^T)$, where $\mtx{d}^{-1}$ is the element-wise inverse of $\mtx{d}$\;
 Compute $\mtx{B}_J \mtx{V}$ and assign it to the corresponding rows of $\mtx{R}$ \tcp*{pilot sampling reutilization}
 Return the matrix $\mtx{R}$ as the ultimate output of this algorithm\;
 \caption{Skeinformer.  }
\label{Alg:incomplete}
\end{algorithm*}

\subsection{Problem formulation}

Given an input sequence $\mtx{X} \in \mb R^{n \times d_{input}}$, where $n$ is the sequence length and $d_{input}$ is the embedding dimension, 
the dot-product attention for a single attention head in transformer~\cite{DBLP:conf/nips/VaswaniSPUJGKP17} is defined as
\begin{equation}
    \nonumber
    \text{Attention}(\mtx{Q},\mtx{K},\mtx{V}) = \text{softmax}\left(\frac{\mtx{QK}^{T}}{\sqrt{p}}\right)\mtx{V} 
    = \mtx{D}^{-1} \mtx{A}\mtx{V}
\end{equation}
where $\mtx{Q} = \mtx{X} \mtx{W}_Q$, $\mtx{K} = \mtx{X} \mtx{W}_K$, and $\mtx{V} = \mtx{X} \mtx{W}_V$.
$\mtx{W}_Q$, $\mtx{W}_K$, $\mtx{W}_V \in \mb R^{d_{input} \times p}$ are the query, key, and value weight metrics that linearly project the input $\mtx{X}$ of dimension $d_{input}$ to an output tensor of dimension $p$.

To ease the future analysis, the softmax term can be rewritten into $\mtx{D}^{-1} \mtx{A}$,
where $\mtx{A} \defeq \exp(\textbf{QK}^{T} / \sqrt{p})$, and $\mtx{D}$ is a diagonal matrix whose diagonal is $\exp(\mtx{Q} \mtx{K}^{T} / \sqrt{p}) \cdot \mtx{1}$ 
($\mtx{1}$ is a size-$n$ vector with all elements being $1$).
% (by convention $\mtx{1}$ is a size-$n$ vector with all elements being $1$).

% \begin{equation}
%     \nonumber
%     \text{Attention}(\textbf{Q},\textbf{K},\textbf{V}) = \mtx{D}^{-1} \mtx{A}\textbf{V}
% \end{equation}

\subsection{Sketching methods}

Beyond current attempts to accelerate self-attention, research in the random matrix approximation community can be potentially applied to fast attention.
Among the theoretical frameworks, the sketching method \citep{woodruff2014sketching} is the most comprehensive one as its general concept can incorporate many different approaches.

The core idea of the sketching method is to replace an original matrix $\mtx{B} \in \mb R^{n_B \times n}$ with its random sketch $\mtx{B} \mtx{S}$,
where $\mtx{S} \in \mb R^{n \times d}$ is a random sketching matrix.
In practice, to apply the sketching method we plug an identity matrix into the original expression,
and then formally replace the identity matrix with the product $\mtx{S} \mtx{S}^T$, 
as the distribution of $\mtx{S}$ is usually designed to satisfy the constraint that 
\begin{align}
\label{Eqn:constraint}
\Expect(\mtx{S} \mtx{S}^T) = \mtx{I}.
\end{align}
% The sketching framework is general, 

Common methods to construct a sketching matrix include sub-Gaussian maps \citep{vershynin2010introduction, DBLP:journals/siamrev/HalkoMT11}, subsampled randomized Hadamard transform (SRHT) \citep{ailon2006approximate, lu2013faster, yang2017randomized}, 
sparse oblivious subspace embeddings \citep{DBLP:conf/icalp/CohenNW16}, very sparse random projection \citep{li2006very}, accumulative sketching \citep{chen2021accumulations}, and sub-sampling sketching (Monte Carlo algorithms) \citep{drineas2006fast}.
Specifically, Informer and Linformer, two efficient transformer-based methods mentioned above, can be understood as applications of sub-sampling sketching and sub-Gaussian maps, respectively.
We further elaborate the connections in the next subsection.

\subsection{Sketching in self-attention approximation}
\label{sec:sketching}

A na\"ive step in applying sketching method to approximate the self-attention output $\mtx{D}^{-1} \mtx{A} \mtx{V}$ is to construct a random sketch of the un-normalized attention score matrix $\mtx{A}$, the bottleneck in computation.
Informer and Linformer construct two types of sketches, $\mtx{A}^T \mtx{S}$ and $\mtx{A} \mtx{S}$ respectively.

% To apply sketching method to the approximation of the self-attention output $\mtx{D}^{-1} \mtx{A} \mtx{V}$, we need to first construct a random sketch of the attention score matrix $\mtx{A}$, the bottleneck in computation.
% Two kind of sketches, $\mtx{A}^T \mtx{S}$ and $\mtx{A} \mtx{S}$, lead to two different efficient transformers, respectively Informer and Linformer.

% In Informer, the sketch $\mtx{A}^T \mtx{S}$ is related to a sub-sampling matrix defined as follows:

\subsubsection{Informer}

Informer selects $d$ important rows of $\mtx{D}^{-1} \mtx{A}$, though deterministically, to represent $\mtx{D}^{-1} \mtx{A}$.
This process can be related to a sketched approximation $\mtx{D}^{-1} \mtx{S} \mtx{S}^T \mtx{A}$, 
% where $\mtx{S}$ is a sub-sampling matrix defined in Definition~\ref{def:subsampling}.
where $\mtx{S}$ is a sub-sampling matrix defined as follows:
% ---------------------------------------------------------
\begin{definition}[Sub-sampling matrix]
\label{def:subsampling}
Consider a discrete distribution which draws $i$ with probability $p_i >0, \forall i \in [n]$.
For a random matrix $\mtx{S} \in \mb R^{n \times d}$, if $\mtx{S}$ has independent and identically distributed (i.i.d.) columns and each column $\mtx{S}^{(j)}$ is
$\frac{1}{\sqrt{d p_i}} \mtx{e}_i$ with probability $p_i$,
where $\mtx{e}_i$ is the $i$-th column of the $n$-by-$n$ identity matrix $\mtx{I}_n$,
then $\mtx{S}$ is called a sub-sampling matrix with sub-sampling probabilities $\{p_i\}_{i=1}^n$.
\end{definition}
% ---------------------------------------------------------

% We will illustrate the connection between the two criteria used to select rows by Informer and sub-sampling sketching in the next two paragraphs.

Some researchers in the field of approximate matrix multiplication have provided a practical guideline for the choice of the sub-sampling probabilities $\{p_i\}_{i=1}^n$ in $\mtx{S}$.
Specifically for the matrix multiplication $\mtx{B} \mtx{C}$ of two arbitrary matrices $\mtx{B}$ and $\mtx{C}$, \citet{drineas2006fast} approximate it with $\mtx{B} \mtx{S} \mtx{S}^T \mtx{C}$ and set the sampling probability $p_i$ in $\mtx{S}$ proportional to the product $\|\mtx{B}^{(i)}\|_2 \|\mtx{C}_{(i)}\|_2$, 
where $\mtx{B}^{(i)}$ is the $i$-th column in matrix $\mtx{B}$ and $\mtx{C}_{(i)}$ is the $i$-th row in matrix $\mtx{C}$.
For the product $\mtx{D}^{-1} \mtx{A}$, the probability in sketching will be $p_i = \frac{\sqrt{\sum_{j=1}^n a_{ij}^2}}{\sum_{j=1}^n a_{ij}}$, 
where $a_{ij}$ is the $j$-th element of the $i$-th row in matrix $\mtx{A}$.

The above sampling probability choice $\{p_i\}_{i=1}^n$ is highly related to the sparsity measurement used in Informer, which is $M_i = \ln{\frac{\sum_{j=1}^n a_{ij}}{(\prod_{j=1}^n a_{ij})^{1/n}}}$.
Here $p_i$ is the ratio between the quadratic mean and the arithmetic mean of $\{a_{ij}\}_{j=1}^n$;
$M_i$ is the logarithm of the ratio between the arithmetic mean and the geometric mean.
It is clear that $M_i$ will increase with $p_i$ as these two ratios will both be large when $\{a_{ij}\}_{j=1}^n$ are highly non-uniform.
We conclude that in Informer, the main idea to select the rows with high sparsity measurement can be taken as a special variant of the sub-sampling method above with probabilities $\{p_i\}$.

\subsubsection{Linformer}
Another type of sketch $\mtx{A} \mtx{S}$ is mentioned (but not finally used) in Linformer. The sketching matrix $\mtx{S}$ takes a form different from sub-sampling. 
The construction of $\mtx{S}$ in Linformer is motivated by Johnson-Lindenstrauss (JL) transform, 
which applies the sketching matrix $\mtx{S}$ satisfying the $(\varepsilon, \delta)$-JL guarantee:
\begin{definition}[Oblivious Johnson-Lindenstrauss guarantee \citep{johnson1984extensions}]
\label{def:JL}
A distribution $\m D$ over $\mb R^{n \times d}$ satisfies ``oblivious Johnson-Lindenstrauss guarantee" 
if for some $\varepsilon, \delta \in (0, 1/2)$:
\begin{equation}
\forall \mtx{b} \in \mb{R}^{n}, 
\underset{\mtx{S} \sim \m D}{\mb{P}}\left(\left|\|\mtx{S} \mtx{b}\|_{2}^{2}-\|\mtx{b}\|_{2}^{2}\right|
    > \varepsilon\|\mtx{b}\|_{2}^{2}\right) < \delta.
\end{equation}
\end{definition}
Specifically, a matrix with   i.i.d. Gaussian elements can  meet the above requirement.
It has been proven \citep{johnson1984extensions} that with $d = \m O(\varepsilon^{-2} \log(1/\delta))$, a Gaussian sketching matrix $\mtx{S}$ can satisfy the oblivious $(\varepsilon, \delta)$-JL guarantee.
To extend the conclusion from a single vector $\mtx{b} \in \mb{R}^{n}$ to a matrix $\mtx{B} \in \mb{R}^{n_B \times n}$, the size $d$ still needs to suffer from an additional $\log n_B$ term \citep{vershynin2010introduction},
which matches the bound in sub-sampling sketching \citep[Theorem~1]{drineas2006fast}.

However, the direct use of Gaussian sketching matrix, i.e. the approximation $\mtx{D}^{-1} \mtx{A} \mtx{S} \mtx{S}^T \mtx{V}$ \citep[Eqn.~(5)]{DBLP:journals/corr/abs-2006-04768}, requires the computation of the whole matrix $\mtx{A}$.
To avoid this computational burden, Linformer replaces the form of sketching method with $\text{softmax}\left((\mtx{Q} \mtx{K}^T / \sqrt{p}) \mtx{S}\right) \mtx{S^T} \mtx{V}$, which sacrifices the accuracy for efficiency in some tasks as shown in later experimental results.

% Later in the empirical experiments we observe the design indeed sacrifices the accuracy for efficiency in some tasks.
% Utilizing the Gaussian sketching matrix with strong theoretical guarantee,  tends to approximate self-attention with  and some other variants.

% removed from emnlp version
% In the next section, we will propose our methods based on the sketching methods above, and validate the improvement upon Informer and Linformer regarding matrix approximation.

\section{Methodology: Skeinformer}
\label{sec:method}

\begin{figure*}[htbp]
\centering
\includegraphics[width=0.75\linewidth]{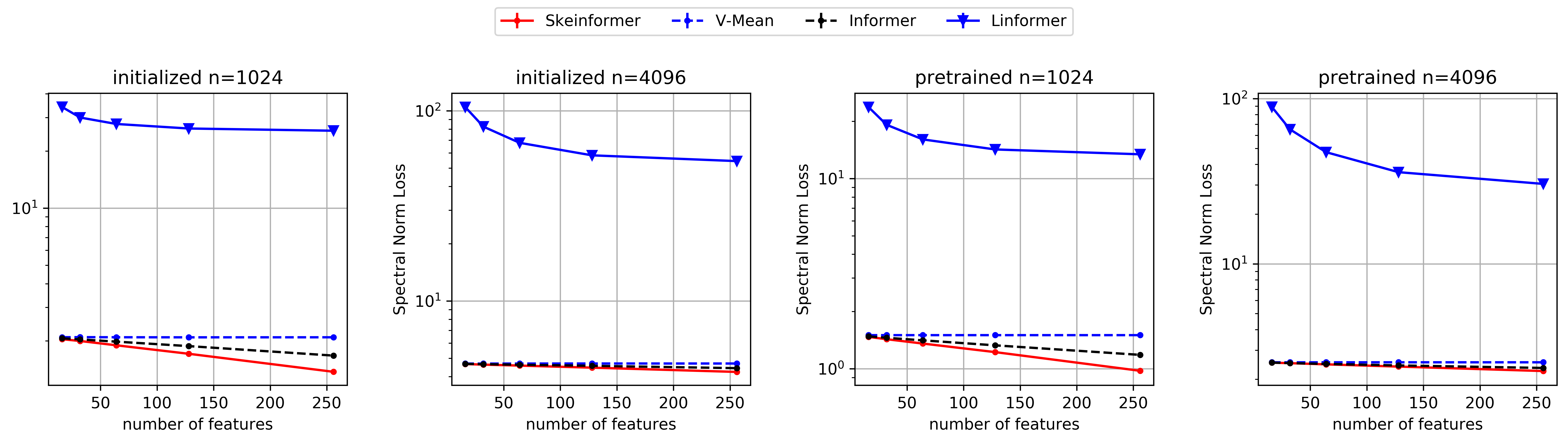}
\caption{
Spectral norm results with the sequence length of 1024 and 4096.
% under different $W_Q, W_K, W_V$ settings, either from initialized or pretrained BERT models. 
Y axis: Lower percentage score means better approximation. 
X axis: Higher number of feature means larger computation cost.
}
\label{fig:norm}
\end{figure*}

Based on the previous discussion, we observe that
Informer omits the information from the value matrix $\mtx{V}$, 
and Linformer deviates from the usual sketching form for efficiency.
To address these issues and fully exploit the power of sketching, we strengthen the attention approximation with the following components.

In Section~\ref{sec:col_samp}, we introduce column importance sampling, which allows the information incorporation from $\mtx{V}$ and accelerate the matrix multiplication (compared to JL transform);
in Section~\ref{sec:rn} and Section~\ref{sec:reutilize}, 
we leverage the sampled columns to perform the row normalization and reuse the pilot row sampling, 
which further improves the approximation and makes the training more stable.

We describe the proposed method \textbf{Skeinformer} in Algorithm~\ref{Alg:incomplete} and verify its performance on matrix approximation in Section~\ref{sec:norm}.
We also provide complexity analysis in Section~\ref{sec:cplx_analysis} to show that our method enjoys the same $\m O(n \log n)$ complexity as other methods.

\subsection{Column sampling}
\label{sec:col_samp}

% For Informer, we have derived the Prob-sparse row selecting is a special variant of the sketching method.
The row selection in Informer has been derived as a special variant of the sketching method and can be further improved by utilizing the information from $\mtx{V}$,
in a form similar to Linformer:
\begin{align*}
\mtx{D}^{-1} \mtx{A} \mtx{S} \mtx{S}^T \mtx{V},
\end{align*}
where $\mtx{S}$ above is a sub-sampling matrix defined in Definition~\ref{def:subsampling} with sampling probabilities
\begin{align*}
p_i \propto \|(\mtx{D}^{-1} \mtx{A})^{(i)}\|_2 \|\mtx{V}_{(i)}\|_2, \quad i=1, 2,\dots, n.
\end{align*}
We remark that using the sub-sampling sketching in this way can both circumvent the computation burden of Gaussian sketching,
and also allow the incorporation of the information from $\mtx{V}$.

As $\mtx{S}$ formally samples some columns from $\mtx{D}^{-1} \mtx{A}$, we name the procedure as column sampling in our method.
The performance regarding the Frobenius norm loss of the approximate matrix multiplication can be guaranteed by the following proposition: 
% -------------------------------------
\begin{prop}[Adapted from Theorem~1 \citep{drineas2006fast}]
\label{prop:Fnorm}
% \vicki{previously B means arbitrary matrix and A means attention socre matrix}
Suppose the attention score matrix $\mtx{B} \defeq \mtx{D}^{-1} \mtx{A} \in \mb R^{n \times n}$, 
the value matrix $\mtx{V} \in \mb R^{n \times p}$, the number of sampled columns $d \in \mb Z^+$ such that $1 \leq d \leq n$, 
and the sampling probabilities $\{p_i\}^n_{i=1}$ are such that $\sum^n_{i=1} p_i = 1$ and such that for a quality coefficient $\beta \in (0,1]$
\begin{align}
\label{eqn:beta_prob}
p_i \geq \beta \frac{\|B^{(i)}\| \|V_{(i)}\|}{\sum_{i'=1}^n \|B^{(i')}\| \|V_{(i')}\|}, \forall i \in [n].
\end{align}
Construct a sub-sampling matrix $\mtx{S} \in \mb R^{n \times d}$ with sub-sampling probabilities $\{p_i\}_{i=1}^n$ as in Definition~\ref{def:subsampling}, and let $\mtx{B} \mtx{S} \mtx{S}^T \mtx{V}$ be an approximation to $\mtx{B} \mtx{V}$.
Let $\delta \in (0, 1)$ and $\eta = 1 + \sqrt{(8/\beta) \log(1/\delta)}$.
Then with probability at least $1-\delta$,
\begin{align}
\label{eqn:guarantee}
\|\mtx{B} \mtx{V} - \mtx{B} \mtx{S} \mtx{S}^T \mtx{V}\|_F^2 \leq \frac{\eta^2}{\beta d} \|\mtx{B}\|_F^2 \|\mtx{V}\|_F^2.
\end{align}
\end{prop}
% -------------------------------------
\noindent \textbf{Remark.}
Proposition~\ref{prop:Fnorm} guides Informer and our method to pick the important rows and columns of the attention score matrix $\mtx{B}$.
In self-attention, it is feasible to compute the norm $\|V_{(i')}\|$ of each row in $\mtx{V}$ with $\m O(n)$ time,
assuming the dimension $p$ in each head is fixed and independent of $n$.
However, similar to Informer, it is inefficient to exactly compute the $\ell$-2 norm of each column in the $n$-by-$n$ matrix $\mtx{B}$,
and we need pilot sampling as well to estimate the norm of the columns in $\mtx{B}$.
We show that $\m O(\log n)$ samples in the pilot sampling is sufficient to guarantee the quality coefficient $\beta \geq \sqrt{\frac13}$ with high probability by the following lemma. 
(The proof is deferred to Appendix~\ref{sec:lem_pilot}.)

% -------------------------------------
\begin{lem}
\label{lem:pilot}
Assume for any $i \in [n]$, $\|\mtx{B}^{(i)}\|^2 / n$ is uniformly lower bounded by a constant $C$, 
where $\mtx{B} \defeq \mtx{D}^{-1} \mtx{A}$.
For another constant $\delta \in (0, 1/2)$, we uniformly sample $d$ indices $\{j_k\}_{k=1}^d$ from $[n]$ with replacement, 
and let $d$ be a constant multiple of $\log(n/\delta)$.
Then with probability at least $1-\delta$, the estimated sub-sampling probabilities
\begin{align}
\label{eqn:est_prob}
\hat p_i \defeq \frac{(\sum_{k=1}^d b_{j_k i}^2)^{\frac12} \|V_{(i)}\|}
{\sum_{i'=1}^n (\sum_{k=1}^d b_{j_k i'}^2)^{\frac12} \|V_{(i')}\|}, \forall i \in [n],
\end{align}
satisfy the constraints~(\ref{eqn:beta_prob}) with $\beta = \sqrt{\frac13}$,
where $b_{ji}$ is the element of $\mtx{B}$ from the $j$-th row and $i$-th column. 
\end{lem}
This lemma states the sub-sampling weights used in our proposed method.
Its computation only requires accesses to $\{\mtx{B}^{(j_k)}\}_{k=1}^d$ obtained from the pilot sampling,
and thus has greatly reduced the time cost.
Combining the preceding lemma and Proposition~\ref{prop:Fnorm}, we conclude that with the sampling probabilities $\{\hat p_i\}^n_{i=1}$ estimated by $\m O(\log n)$ pilot samples, 
the sampled $d$ important columns suffice to capture the essence of the original output $\mtx{B} \mtx{V}$.
We close this subsection with a remark that the theoretical result that sub-sampling sketching can well approximate the original self-attention, 
indeed matches the rank collapse phenomenon observed by \citet{dong2021attention} that self-attention can be well approximated by a low-rank matrix.

% TODO: assume the variance for each column is bdd, and use max subgaussian to prove that the estimated sum is within [0.5 SS, 1.5 SS] (sum of squares) whp, then at worst beta = sqrt(0.5/1.5)

% The proposition above also justifies the pilot sampling in Informer with $\m O(\ln n)$ samples. 

\subsection{Adaptive row normalization}
\label{sec:rn}

% Rather than directly applying the sketching method, our proposed method also includes a trick of adaptive row normalization.
% Another important observation from the experiment on spectral norm loss is the trick 

In addition to the theoretical guarantee of the sub-sampling sketching method,
we identify an important component behind Informer, row normalization, 
which implicitly fills the un-selected rows with $\frac1n$.
Toy experiments in Section~\ref{sec:norm} reveal that even the rank-one pure row normalization benchmark $\frac1n \mtx{1} \mtx{1}^T V$, as an ablation, will have acceptable spectral norm loss $\|\mtx{D}^{-1} \mtx{A} \mtx{V} - \frac1n \mtx{1} \mtx{1}^T V\|_2$.
Therefore, we incorporate adaptive row normalization to provide an even better attention score distribution in each row. 
It fills un-selected columns with the averaged selected columns.
% where $\mtx{1}$ is a length $n$ vector with all elements $1$. 
% Given the prior knowledge that row normalization helps, besides the rows selected by sketching part, 
% we also add a row normalization trick to our method.
Moreover, from the model training perspective, it allows the whole value matrix $\mtx{V}$ in Skeinformer to participate in the computation (compared to only using the sub-sampling sketch $\mtx{S}^T \mtx{V}$), 
and thus can improve the efficiency of updating $\mtx{W}_V$ during the training.

% Based on the observation above, besides the important columns sampled, 
% we add a row normalization trick to our method to provide a better attention score distribution in each row.
Specifically, in adaptive row normalization any row in the matrix $\mtx{A}$ can be divided into two parts,
the exactly computed elements in the selected columns with indices $\{j'_k\}_{k=1}^d \subset [n]$
and the other elements in the un-selected columns.
For the latter, in each row, we set all the un-selected elements as the geometric mean of the selected ones,
considering the exponentiation in softmax.
We then perform row normalization based on the above construction, 
in which the $i$-th diagonal element in $\mtx{D}$ is estimated as
\begin{align}
\label{eqn:est_dii}
\hat d_{ii} = \sum_{k=1}^d a_{i j_k} + (n-d) (\prod_{k=1}^d a_{i j_k})^{\frac1d},
\end{align}
where each $a_{ij}$ is the corresponding element in matrix $\mtx{A}$.
Next we normalize rows composed of exact elements in the selected columns, and the other elements estimated with the mean value above.
We comment that though the component of adaptive row normalization makes the proposed method inapplicable to Proposition~\ref{prop:Fnorm},
it benefits the performance on matrix approximation and avoid the cost to compute the diagonal normalization matrix $\mtx{D}$. (c.f. Section~\ref{sec:norm})

\subsection{Pilot sampling reutilization}
\label{sec:reutilize}

% The third component in Skeinformer is the reutilization of pilot sampling in Lemma~\ref{lem:pilot} to improve the matrix approximation.

Since we have already computed $\mtx{B}_J$ in pilot sampling step (defined in Ln.~3 of Algorithm~\ref{Alg:incomplete}), 
we can exactly reproduce the $d$ rows in the original self-attention output with an additional product $\mtx{B}_J \mtx{V}$ in $\m O(n \log n)$ time.
This allows for more precise approximation with little cost. 
In addition, the computation of those rows involves the whole key matrix $\mtx{K}$, which benefits the training of the parameters $\mtx{W}_K$.

% Adding the trick to Algorithm~\ref{Alg:incomplete} leads to a variant Skeinformer+, which is described in Algorithm~\ref{Alg:skein}.

\subsection{Implementation Details}
\label{sec:padding}
%  and adaption to masks in NLP tasks
Applying the sub-sampling-based methods requires the support for padding masks commonly used in Natural Language Processing tasks.
However, a na\"ive implementation of Algorithm~\ref{Alg:incomplete} will result in the unnecessary sampling of the padding tokens. 
Therefore, we count the number of the unpadded tokens $m$, and only perform the pilot sampling within the certain range $[m]$.
After the matrix $\mtx{B}_J$ is computed, we set its columns belonging to the padded part to be all zero,
so that the probability $\hat p_i$ of choosing column $i$ from the padded part will be zero and the column will not be sampled in the later importance sampling.
Similar modifications can also be applied to Informer to enable its applications in NLP tasks in Section~\ref{sec:experiment}.

% \william{We validate the benefits of this step in Section~\ref{sec:experiment} through experiments.}

\iffalse
use pilot sampling at the end as well to update all parameters (Q,K)
Finally, we remark Informer doesn't fully utilize the pilot computation of elements in $\mtx{A}$ for row choosing.
In the complete Skeinformer, we plug the initially sampled rows into the approximation for matrix $\mtx{D}^{-1} \mtx{A}$.
support for the padding mask
\fi

\subsection{Complexity analysis}
\label{sec:cplx_analysis}

With the mild assumption in Lemma~\ref{lem:pilot}, we claim that our method can have an $\m O(n \log n)$ time and space complexity.
The claim is shown by the following analysis of the complexity, which heavily relies on the notations in Algorithm~\ref{Alg:incomplete}.

First, we point out that the row/column retrieving operation after index sampling can be implemented by only forming a view and thus the cost is negligible.
For Line~$1\sim4$ in Algorithm~\ref{Alg:incomplete}, the time complexity of the uniform pilot sampling is $\m O(d) = \m O(\log n)$,
while the computation of the matrix $\mtx{B}_J$ and the corresponding probabilities $\{\hat p_i\}^n_{i=1}$ costs $\m O(nd) = \m O(n \log n)$ time and space.
For Lines~$5\sim7$, with probabilities $\{\hat p_i\}^n_{i=1}$, the importance sampling takes $\m O(n + d \log n) = \m O(n)$ time, 
and similar to the computation above it takes $\m O(n \log n)$ time and space to obtain $\mtx{A}_{J'}$ and $\mtx{R}_{J'}$.
For Lines~$8\sim10$, it is clear that the three vectors $\mtx{g}$, $\mtx{d}$, and $\mtx{v}$ can be computed in $\m O(n \log n)$ time.
As for the last step in Line~$11$, since it just requires the matrix product involving a diagonal matrix, we can finish the computation also in $\m O(n \log n)$ time and space.
In summary, the total time and space complexity for Algorithm~\ref{Alg:incomplete} is at most $\m O(n \log n)$, much lower than the $\m O(n^2)$ complexity for the original softmax self-attention.

\noindent \textbf{Remark.}
The complexity above is derived based on the high probability bound in Proposition~\ref{prop:Fnorm},
% theoretical guarantee that Inequality~(\ref{eqn:guarantee}) will hold with high probability, 
which is different than the derivation by some previous methods to claim the linear $\m O(n)$ complexity.

%%%%%%%%%%%%%%%%%%%%%%%%%%%%%%%%%%%%%%%
%%%%%%%%%%%%%%%%%%%%%%%%%%%%%%%%%%%%%%%

\section{Approximation Evaluation}
\label{sec:norm}

\begin{table*}[h]
\small
\centering

\begin{tabular}{l|c|c|c|c|c|c}
\hline
Models & Text & ListOps & Retrieval  & Pathfinder  & Image  & Average \\
\hline
Standard~\cite{DBLP:conf/nips/VaswaniSPUJGKP17} & 57.69		 & 		38.15		 & 		80.10		 & 		73.59	 & 			37.97		 & 		57.50\\
$\cdot$ w/o dropout & 59.44		 & 		38.17		 & 		79.35	 & 			72.35		 & 		37.58		 & 		57.38\\
\hline
V-mean & 65.29	 & 			28.78		 & 		80.49		 & 		61.01	 & 			34.33	 & 			53.98\\
BigBird~\cite{DBLP:conf/nips/ZaheerGDAAOPRWY20} &  61.91	 & 			38.86	 & 			79.73	 & 			71.75	 & 			35.00		 & 		57.45\\
Performer~\cite{DBLP:journals/corr/abs-2009-14794}& 57.67	 & 			37.70	 & 			75.69	 & 			56.50	 & 			37.40		 & 		52.99 \\
Nystromformer~\cite{DBLP:journals/corr/abs-2102-03902} &  60.91		 & 			37.76	 & 				79.87	 & 				72.53	 & 				31.93	 & 				56.60\\
Reformer~\cite{DBLP:conf/iclr/KitaevKL20} &  62.69	 & 			37.94	 & 			78.85	 & 			69.21	 & 			36.42		 & 		57.02\\
\hline
Linformer~\cite{DBLP:journals/corr/abs-2006-04768} &  58.52	 & 				37.97	 & 				77.40 & 					55.57	 & 				37.48	 & 				53.39\\
$\cdot$ w/ unreduced JLT &  59.12	& 				37.48		& 			79.39	& 				68.45	& 				35.96	& 				56.08\\
Informer~\cite{DBLP:journals/corr/abs-2012-07436} &  61.55	 &			38.43	 &			80.88	 &			59.34		 &		36.55		 &		55.35 \\
$\cdot$ w/ padding mask &  60.98		& 			37.26		& 			79.92		& 			62.51			& 		37.19		& 			55.57\\
\hline
\textbf{Skeinformer} &  62.47		& 			38.73		& 			80.42		& 			71.51	& 				37.27		& 			\textbf{58.08}\\
$\cdot$ w/ uniform sampling &  64.48		&			30.02		&			80.57			&		64.35		&			36.97		&			55.28\\
$\cdot$ w/o row normalization &  60.67		&			37.69	&				78.67		&			66.35			&		37.06			&		56.09\\
$\cdot$ w/ simple row normalization & 60.26		&			38.35		&			78.97		&			65.41		&			39.72			&		56.54\\
$\cdot$ w/o  pilot sampling reutilization & 62.39		&			38.12		&			79.88		&			71.53			&		37.20		&			57.83\\
\hline
\end{tabular}
\caption{
Classification accuracy (\%) on the test sets of LRA benchmark. 
}
\label{table:lra_acc_aaai}
\end{table*}

As a preliminary justification of our proposed methods, we compute the spectral norm loss, a common metric for approximate matrix multiplication, to evaluate the effect of different models to approximate the original self-attention.
We compare the spectral norms of the differences between the outputs from vanilla self-attention and other fast attention methods given the same input $\textbf{Q}, \textbf{K}, \textbf{V}$.
% $\left\| \textbf{O}(\textbf{Q}, \textbf{K}, \textbf{V}) - \text{softmax}(\textbf{QK}^{T} / \sqrt{p})  \right\|_2  / \left\| \text{softmax}(\textbf{QK}^{T} / \sqrt{p}) \right\|_2$
Specifically we compute $\|\mtx{B} \mtx{V} - \mtx{R}\|_2$, where $\mtx{B} \defeq \mtx{D}^{-1} \mtx{A}$ is the attention score matrix in the original method, and $\mtx{R}$ is the output of each approximation method.

The inputs $\textbf{Q}, \textbf{K}, \textbf{V}$ are constructed in the following process:
We first truncate the raw text of sequence length $1024$ or $4096$ from Wikitext-2 dataset~\citep{DBLP:conf/iclr/MerityX0S17} into sequences of length $512$
% tokenize them with a pretrained Tokenizer,
and embed the tokenized input with the embedding layer in a pretrained bert-base-cased model using Huggingface's implementation~\citep{DBLP:journals/corr/abs-1910-03771}.
Then we transform the input $X$ into $\textbf{Q}, \textbf{K}, \textbf{V}$ with the query, key, and value weight metrices from a pretrained model or a randomly initiated model. 

% The weight matrices in two settings transform input $X$ into different distributions.

We report the spectral norm loss of different sketching-based methods in Figure~\ref{fig:norm}.
% (The result for Linformer is deferred to Appendix~\ref{sec:norm_eval}, as it falls in a different range and influences the differentiation of other curves.)
The results are averaged over $768$ trials, and the error bars in  the figure represent the standard error of the reported values.
For size $d$ ($x$-axis) in the simulation, either the number of columns/rows selected or the projection dimension, it is set in the range from $2^3$ to $2^8$. 
% More features generally require more computation resources and have better approximation quality.

A special baseline ``V-Mean" always uses a rank-one matrix $\frac1n \mtx{1} \mtx{1}^T V$ to approximate the original self-attention, and thus its approximation error does not change with the size $d$.
``V-Mean" can also be seen as an ablation for the row normalization step (equivalent to adaptive row normalization without any sub-samples).
From the results we observe the row normalization step greatly contributes to the approximation of self-attention that involving a softmax structure.
Among the candidates, Skeinformer tends to have the smallest spectral norm loss, especially when $d$ is large.
We conclude that our algorithm design attains a higher accuracy than Informer and Linformer (shown in Figure~\ref{fig:norm}), regarding the performance of matrix approximation.

% We can also observe that Linformer indeed provides a result totally distinct from the original method due to the violation of the sketching form,
% while our methods attain a high accuracy and thus effectively approximate the original self-attention.
% \william{add the results of $d=1$ setting to show performer have a much worse performance than ours.}

% We compare the results with different sequence lengths and different number of features used in attention approximation methods. 
% We set the number of features in the range of $2^3$ to $2^8$. More features usually require more computation resources.

% For the number of features $d$ in the x-axis, we use this quantity to control the number of elements used by different methods for approximating $\mtx{B}$ to all be $nd$,  so as to ensure the comparison is as fair as possible.

% The data used in each experiment is illustrated in the caption of Figure~\ref{fig:norm}.

\section{Experiment}
\label{sec:experiment}

\begin{table*}[h]
\small
\centering

\begin{tabular}{l|c|c|c|c|c|c|c|c|c|c|c|c|c|c|c}
\hline
Models & \multicolumn{3}{c|}{Text} & \multicolumn{3}{c|}{ListOps} & \multicolumn{3}{c|}{Retrieval}  & \multicolumn{3}{c|}{Pathfinder} & \multicolumn{3}{c}{Image} \\
 & step  & time & accu  & step  & time & accu   & step  & time & accu    & step  & time & accu  & step  & time & accu  \\ 
\hline
Standard & 11 & 	51 & 	8 & 		9 & 	22 & 	4	 & 	20 & 	53 & 	4	 & 	21 & 	14 & 	4	 & 	4 & 	21 & 	4\\
$\cdot$ w/o dropout &  6 & 	39 & 	16 & 		6 & 	20 & 	8 & 		16 & 	42 & 	8	 & 	18 & 	12 & 	8	 & 	4 & 	15 & 	8\\
\hline
V-mean & 8 & 	4 & 	1	 & 	10 & 	4 & 	1	 & 	27 & 	4 & 	1 & 		16 & 	4 & 	1 & 		7 & 	4 & 	1\\
BigBird & 6	 & 21 & 	2 & 		9 & 	17 & 	2 & 		15 & 	22 & 	2 & 		17 & 	18 & 	2 & 		3 & 	19 & 	1\\
Performer & 30 & 	3 & 	2 & 		10 & 	9 & 	2 & 		13 & 	12 & 	2	 & 	9 & 	10 & 	2 & 		5 & 	9 & 	1\\
Nystromformer & 12 & 	12 & 	2 & 		10 & 	12 & 	2	 & 	17 & 	13 & 	2 & 		23 & 	20 & 	4 & 		3 & 	10 & 	1\\
Reformer & 7 & 	11	 & 2	 & 	14 & 	8 & 	2	 & 	26 & 	11 & 	1 & 		24 & 	9 & 	2 & 		5 & 	12 & 	1\\
\hline
Linformer & 7 & 	8 & 	2 & 		11 & 	6 & 	2 & 		23 & 	8 & 	1 & 		7 & 	7 & 	2	 & 	4 & 	7 & 	1\\
$\cdot$ w/ unreduced JLT  & 6 & 	37 & 	16 & 		4 & 	21 & 	8 & 		24 & 	36 & 	16	 & 	12 & 	15 & 	4 & 		4 & 	22 & 	2 \\
Informer & 8 & 	33 & 	8 & 		10 & 	22 & 	8 & 		25 & 	37 & 	4	 & 	9 & 	26 & 	8	 & 	7 & 	25 & 	2\\
$\cdot$ w/ mask & 7 & 	26 & 	4 & 		4 & 	22 & 	4 & 		31 & 	36	 & 2 & 		7 & 	16 & 	4 & 		5 & 	23 & 	2\\
\hline
\textbf{Skeinformer}  & 6 & 	10 & 	2	 & 	7 & 	10 & 	2	 & 	20 & 	11 & 	1 & 		18 & 	9 & 	2 & 		6 & 	12	 & 1\\
$\cdot$ w/ US & 8 & 	8 & 	1	 & 	3 & 	7 & 	1 & 		26 & 	7 & 	1 & 		19 & 	7 & 	1 & 		8 & 	8 & 	1\\
$\cdot$ w/o RN & 6 & 	25 & 	8 & 		6 & 	16 & 	4	 & 	14 & 	56	 & 16 & 		13 & 	11 & 	2	 & 	5 & 	16 & 	2\\
$\cdot$ w/ SRN & 7	 & 7 & 	1	 & 	6 & 	8 & 	1 & 		19 & 	8 & 	1	 & 	14 & 	7 & 	1 & 		6 & 	11 & 	1\\
$\cdot$ w/o PSR & 7	 & 7 & 	1	 & 	6 & 	7 & 	1 & 		22	 & 9	 & 1	 & 	17 & 	7 & 	1	 & 	6	 & 10 & 	1\\
\hline
\end{tabular}
\caption{
Training steps (k), training time (minute) per thousand steps, and batch accumulation steps on LRA benchmark.  Smaller training steps implies faster convergence. 
Less training time per thousand steps indicates higher time efficiency.
Less batch accumulation steps means larger actual batch size and higher space efficiency.
\textbf{US}: Uniform Sampling. 
\textbf{RN}: Row Normalization.
\textbf{SRN}: Simple Row Normalization .
\textbf{PSR}: Pilot Sampling Reutilization.
}
\label{table:lra_timespace_aaai}
\end{table*}

\subsection{Setup}

% To further justify the effectiveness of the proposed Skeinformer method, 
We compare our method with the standard quadratic self-attention \cite{DBLP:conf/nips/VaswaniSPUJGKP17}, Big Bird \cite{DBLP:conf/nips/ZaheerGDAAOPRWY20}, Linformer \cite{DBLP:journals/corr/abs-2006-04768}, Informer \cite{DBLP:journals/corr/abs-2012-07436}, Performer \cite{DBLP:journals/corr/abs-2009-14794}, and Nystr\"omformer \cite{DBLP:journals/corr/abs-2102-03902}.
In addition to their vanilla implementations, we compare with standard self-attention without dropout (since most fast attention methods do not have this component), Linformer with unreduced Johnson-Lindenstrauss Transform (the original form that Linformer deviates from), and Informer with padding masks (as introduced in \ref{sec:padding}). 
``V-Mean" always uses a rank-one matrix $\frac1n \mtx{1} \mtx{1}^T V$ to approximate the original self-attention, and serves as a special baseline against sampling.

To justify the effectiveness of the proposed components, we perform ablation studies, including replacing the column importance sampling with uniform sampling, disabling the adaptive row normalization or replacing it with the simple row normalization implemented in Informer, and disabling the pilot sampling reutilization.

Since some methods cannot be analyzed under the sketching framework, and their complexity is derived under different theoretical settings,
we compare their practical performance through experiments on Long Range Arena (LRA) benchmark~\cite{DBLP:journals/corr/abs-2011-04006}.
LRA is designed for long-context scenarios, and will be more appropriate for evaluating efficient transformers for long sequences than more general Natural Language Understanding tasks with much shorter input context.
We evaluate on five classification tasks in LRA benchmark: ListOps \cite{DBLP:conf/naacl/NangiaB18}, Text Classification on IMDb review dataset \cite{DBLP:conf/acl/MaasDPHNP11}, Document Retrieval on AAN dataset \cite{DBLP:journals/lre/RadevMQA13},  Pathfinder \cite{DBLP:conf/nips/LinsleyKVWS18}, and Image Classification~\cite{krizhevsky2009learning})

% We report the classification accuracy, training step, and training time per thousand steps for each task.
% The complete settings of the experiments below and more supplementary results can be found in Appendix~\ref{sec:exp}.

\subsection{Implementation details}

As it is not realistic to exhaustively fine-tune all models and search for the best performance under limited computation resources, we instead replace the self-attention module in transformer with the various drop-in attention methods and keep other experimental settings the same.
Following \cite{DBLP:journals/corr/abs-2102-03902} we use a 2-layer transformer model with 64 embedding dimensions, 128 hidden dimensions, and 2 attention heads for all experiments.
Mean pooling is used in all classifiers. 

For comparable computation complexity, we control the number of features used in all methods, 
which leads to $256$ as the number of features in Skeinformer, 
$256$ as $k$ in Linformer,
$256$ as the number of landmarks in Nystr\"omformer, 
$(256/\log n)$ as the factor in Informer, 
and $256$ as the number of features in Performer.
Additionally, the number of random blocks and block size in Big Bird are by default $3$ and $64$, under which setting Big Bird will visit $640\cdot n$ elements in the attention matrix while other models visit $256\cdot n$ elements.
% big bird: n*nbf = (r+w+g)*b*n, block_size = nb_feat / 10
A clearer complexity evaluation on the FLOPs of each method is provided in Appendix~\ref{sec:FLOPs}.

We use Adam optimizer \cite{DBLP:journals/corr/KingmaB14} with a learning rate of $1e-4$.
Batch size is selected conditioned on the memory requirements of Skeinformer, which leads to $128$ for Text Classification, $256$ for ListOps, $64$ for Document Retrieval, $512$ for Pathfinder and $256$ for Image.
For methods reporting out-of-memory errors, we apply gradient accumulation and report the accumulated steps in Appendix.
Instead of setting a fixed epoch number, we train all models until convergence with a stopping strategy (if better performance is not observed for 10 checking steps on the validation set we will stop the training process) and report the required steps and average time also in Appendix. 

We conduct all experiments on one Tesla V100 SXM2 16GB.
For numerical consistency, all experiment results are averaged across three random runs.

\subsection{Results}

% Unlike many trending NLP benchmarks, evaluation on efficient transformers should not be based on classification accuracy alone.
A transformer is considered efficient when it  
(1) reduces space complexity and supports larger sequence length and larger batch size, 
(2) reduces time complexity with less training time per step and less total time to converge, 
and (3) shows comparable performance with vanilla softmax without much loss from approximation.
Accordingly, we conclude the results shown in Table~\ref{table:lra_acc_aaai} and Table~\ref{table:lra_timespace_aaai} with the following observations:

% \textbf{Space Efficiency: Instead of time complexity reduction, the most significant advantage of efficient transformers is less space requirement that enables larger batch size.} 
\textbf{Space Efficiency: Skeinformer requires less space and enables larger batch size.} 
% We show the batch size used during the training for each method in Table~\ref{table:bz} in Appendix.
Within a certain range, a larger batch size usually means more accurate gradient estimations.
Therefore, we simulate the case of real-world applications of efficient transformers that models are trained with their maximum batch size conditioned on memory.
In general, Skeinformer requires the least space during the training process and supports a larger batch size.

\textbf{Time Efficiency: Skeinformer has consistently less time consumption.} 
We compare the steps and total training time required for convergence. 
% Nystr\"omformer or Informer may have higher accuracy on Pathfinder but with the price of more time to converge, even higher than standard method.
Our method takes consistently much less total training time to converge comparing to the vanilla self-attention.
For example, Skeinformer brings $4.78\%$ accuracy improvement and nearly $9\times$ speed-up on text classification over the standard method.

\textbf{Fast Convergence: Skeinformer efficiently converges to the long-time limit.} 
Regarding the training efficiency, we focus on how soon the model can attain the stationary distribution of its long-time limit \citep{DBLP:conf/nips/HeLT19}.
The loss decay plot on ListOps in Figure~\ref{fig:valloss_all} in Appendix~\ref{sec:exp}
% (results for all tasks are shown in Appendix due to space limitations)
shows significant differences in the convergence rate of each method in addition to classification accuracy. 
% Skeinformer and Informer can efficiently converge to the long-time limit, while Linformer fails to attain a stationary distribution as good as the original softmax.

% From the plot, we can conclude that Linformer and our two proposed variants can much more efficiently converges to the long-time limit,while for efficiency Linformer somewhat sacrifices the accuracy so that it fails to attain a stationary distribution as good as the original softmax.
% Figure \ref{fig:valloss} shows the loss changes on validation set of ListOps during training (results for other tasks are shown in Appendix due to space limit).
% From a machine learning practitioner's perspective, we care more about how soon the model can attain the stationary distribution of its long-time limit \citep{DBLP:conf/nips/HeLT19}.

\textbf{Comparable General Performance: Most $\wt{\m O}(n)$ attention acceleration methods have comparable or better performance with vanilla attention.}
% We use a different training strategy with previous work that we allow each method to be trained until convergence by setting a stopping patience.
% We report the accuracy of efficient transformers in each task averaged over three trials in Table~\ref{table:lra}.
After all models converge to their long-time limits,
Linformer tends to have worse performance possibly due to the violation of the sketching form,
while Skeinformer has the best overall performance.

The pitch of most efficient transformers above is to mimic the original transformer with a reduced computation cost.
While surprising, those approximation methods tend to outperform the original transformer in most tasks.
We speculate the reason behind this phenomenon is that a good approximation can recover the main signals in the original self-attention matrix,
and also restrain the noise via the sparse / low-rank structure.
A similar phenomenon can be found in CNN \citep{sanyal2018robustness}, that a low-rank regularizer, such as SVD, applied to the representation of the intermediate layers can allow the model to have lower prediction errors.

% This speculation motivates us to turn to some theoretical framework for matrix approximation to better analyze the fast attention methods, which will potentially benefit transformer pruning, compression and distillation.

\section{Conclusion}
\label{sec:conclusion}

We conclude in this paper that sketching can be applied as a theoretical framework for analyzing fast attention models,
through which we are able to recognize the potential improvements upon previous work.
Theoretical results are provided to guarantee the high accuracy of the approximation to the original self-attention by our proposed Skeinformer.
We also empirically validate the contributions of the components in Skeinformer, including column sampling, adaptive row normalization and pilot sampling reutilization, with extensive comparisons with various baseline methods and ablation methods.
% (2) Most $\wt{\m O}(n)$ attention acceleration methods have comparable or even better performance with vanilla attention, for which we speculate that a good approximation can recover the main signals in the original self-attention matrix and also restrain the noises via the sparse / low-rank structure.

% A direct further development direction is to extend the sketching method from sub-sampling to other promising alternatives,
% such as very sparse random projection \citep{li2006very} and accumulative sketching \citep{chen2021accumulations}.
% It will also be interesting to study how to decide the optimal size $d$ of the sketching matrix $\mtx{S}$ for either efficiency or accuracy of the transformer model.

\section{Acknowledgments}
This research is based upon work in part supported by the Office of the Director of National Intelligence (ODNI), Intelligence Advanced Research Projects Activity (IARPA), via contract No. FA8650-17-C-9116, and U.S. DARPA KAIROS Program No. FA8750-19-2-1004. This work is also in part supported by NSF grant DMS-1810831. The views and conclusions contained herein are those of the authors and should not be interpreted as necessarily representing the official policies, either expressed or implied, of DARPA, ODNI, IARPA, or the U.S. Government. The U.S. Government is authorized to reproduce and distribute reprints for governmental purposes notwithstanding any copyright annotation therein.

\clearpage
\bibliography{main}
\bibliographystyle{aaai22}
\clearpage
\appendix

\section{Further experiment results}
\label{sec:exp}

\subsection{Details of training time}

\begin{table*}[h]
\small
\centering
\begin{tabular}{l|cc|cc|cc|cc|cc}
\hline
Models & \multicolumn{2}{c|}{Text } & \multicolumn{2}{c|}{ListOps } & \multicolumn{2}{c|}{Retrieval (64)}  & \multicolumn{2}{c|}{Pathfinder }  & \multicolumn{2}{c}{Image}  \\
 & steps & time &  steps & time & steps & time & steps & time& steps & time \\
\hline

Standard & 10.67 & 	540.00 & 			8.53 & 	190.29		 & 	19.80 & 	1,054.67 & 			21.40 & 	297.78 & 			4.33 & 	92.73		\\
$\cdot$ w/o dropout & 5.67 & 223.79 & 5.58 & 108.90 & 16.30 & 682.72 & 17.58 & 207.14 & 4.23 & 63.01 \\
\hline
V-mean  & 8.00 & 28.92 & 9.77 & 40.46 & 26.50 & 103.41 & 15.60 & 57.27 & 7.38 & 32.77 \\
BigBird & 6.00 & 123.52 & 9.07 & 156.68 & 14.90 & 323.83 & 17.47 & 311.40 & 3.20 & 60.28 \\
Performer & 30.33 & 79.64 & 9.90 & 92.14 & 12.50 & 156.22 & 8.83 & 91.91 & 5.13 & 45.87 \\
Nystromformer & 11.50 & 140.02 & 9.87 & 121.19 & 17.10 & 228.26 & 22.60 & 442.59 & 2.92 & 30.06 \\
Reformer  & 6.67 & 70.17 & 14.20 & 117.55 & 26.40 & 297.59 & 24.13 & 223.28 & 4.78 & 56.84 \\
\hline
Linformer & 6.50 & 51.38 & 11.43 & 71.48 & 23.20 & 187.51 & 7.30 & 50.35 & 4.25 & 28.27 \\
$\cdot$ w/ unreduced JLT & 6.25 & 230.46 & 3.98 & 85.43 & 23.85 & 856.83 & 12.01 & 182.18 & 4.30 & 94.75 \\
Informer & 8.00 & 265.06 & 10.13 & 221.79 & 24.70 & 901.99 & 9.47 & 247.50 & 6.72 & 167.37 \\
$\cdot$ w/ padding mask & 7.00 & 181.61 & 4.20 & 90.31 & 30.90 & 1,110.98 & 7.02 & 110.83 & 5.10 & 115.14 \\
\hline
\textbf{Skeinformer}  & 6.17 & 59.23 & 7.32 & 70.71 & 20.40 & 216.36 & 17.89 & 165.54 & 5.57 & 66.02 \\
$\cdot$ w/ uniform sampling & 8.00 & 60.79 & 3.20 & 21.30 & 26.10 & 174.96 & 18.72 & 136.05 & 7.67 & 59.52 \\
$\cdot$ w/o row normalization & 6.17 & 154.31 & 5.90 & 94.54 & 13.80 & 768.97 & 13.21 & 146.88 & 4.50 & 69.83 \\
$\cdot$ w/ simple row normalization & 6.67 & 45.32 & 6.27 & 51.12 & 19.30 & 155.07 & 13.62 & 93.19 & 5.78 & 65.17 \\
$\cdot$ w/o  pilot sampling reutilization & 6.50 & 46.47 & 6.25 & 45.68 & 22.20 & 192.69 & 16.64 & 117.99 & 6.23 & 63.51 \\
\hline

\end{tabular}
\caption{Training steps (k) and total training time (minute) on LRA benchmark.
}
\label{table:total_time}
\end{table*}

Table~\ref{table:total_time} reports the total training time for one random run on each task. 
Our method takes consistently much less total training time to converge compared to the vanilla self-attention.

\begin{table*}[h]
\small
\centering
\begin{tabular}{l|cc|cc|cc|cc|cc}
\hline
Models & \multicolumn{2}{c|}{Text (128)} & \multicolumn{2}{c|}{ListOps (256)} & \multicolumn{2}{c|}{Retrieval (64)}  & \multicolumn{2}{c|}{Pathfinder (512)}  & \multicolumn{2}{c}{Image(256)}  \\
 & bz & accu & bz & accu &bz & accu &bz & accu &bz & accu \\
\hline

Standard & 16 & 8 & 64 & 4 & 16 & 4 & 128 & 4 & 64 & 4 \\	
$\cdot$ w/o dropout & 8 & 16 & 32 & 8 & 8 & 8 & 64 & 8 & 32 & 8 \\
\hline
V-mean & 128 & 1 & 256 & 1 & 64 & 1 & 512 & 1 & 256 & 1 \\
BigBird &  64 & 2 & 128 & 2 & 32 & 2 & 256 & 2 & 256 & 1 \\ 
Performer& 64 & 2 & 128 & 2 & 32 & 2 & 256 & 2 & 256 & 1 \\
Nystromformer  & 64 & 2 & 128 & 2 & 32 & 2 & 128 & 4 & 256 & 1 \\
Reformer  & 64 & 2 & 128 & 2 & 64 & 1 & 256 & 2 & 256 & 1 \\
\hline
Linformer  & 64 & 2 & 128 & 2 & 64 & 1 & 256 & 2 & 256 & 1 \\
$\cdot$ w/ unreduced JLT & 8 & 16 & 32 & 8 & 4 & 16 & 128 & 4 & 128 & 2 \\
Informer & 16 & 8 & 32 & 8 & 16 & 4 & 64 & 8 & 128 & 2 \\
$\cdot$ w/ padding mask & 32 & 4 & 64 & 4 & 32 & 2 & 128 & 4 & 128 & 2 \\
\hline
\textbf{Skeinformer}& 64 & 2 & 128 & 2 & 64 & 1 & 256 & 2 & 256 & 1 \\
$\cdot$ w/ uniform sampling & 128 & 1 & 256 & 1 & 64 & 1 & 512 & 1 & 256 & 1 \\
$\cdot$ w/o row normalization& 16 & 8 & 64 & 4 & 4 & 16 & 256 & 2 & 128 & 2 \\
$\cdot$ w/ simple row normalization & 128 & 1 & 256 & 1 & 64 & 1 & 512 & 1 & 256 & 1 \\
$\cdot$ w/o  pilot sampling reutilization & 128 & 1 & 256 & 1 & 64 & 1 & 512 & 1 & 256 & 1 \\
\hline

\end{tabular}
\caption{Actual batch size (bz) during gradient accumulation steps (accu) under constraints of memory usage. 
}
\label{table:batch_size}
\end{table*}

In the experiments, the batch size is selected conditioned on the memory requirements, which leads to $128$ for Text Classification, $256$ for ListOps, $64$ for Document Retrieval, $512$ for Pathfinder, and $256$ for Image.
For methods reporting out-of-memory errors, we apply gradient accumulation of appropriate steps.
We show the actual batch size during gradient accumulation steps in the training process in Table~\ref{table:batch_size}. 
Skeinformer requires less space for training and thus supports the larger batch size comparing to the vanilla self-attention.

\subsection{FLOPs}
\label{sec:FLOPs}

We conclude in this subsection the floating point operations (FLOPs) of each candidate model (excluding Reformer and the ablation models) in Section~\ref{sec:experiment} in the main paper.
To ease the notation, given the sequence length $n$, we fix $p = 32, d=256$. 
Assuming the matrices $\mtx{Q}, \mtx{K}, \mtx{V}$ are given and omitting the non-leading term, we report the FLOPs of each model in Table~\ref{tab:FLOPs}.
We additionally comment that Reformer is excluded from the table because its FLOPs are not fixed and depend on the frequency of collision after hashing of tokens, which changes with the input sequence.
Table~\ref{table:lra_timespace_aaai} in the main paper empirically shows its average runtime per one thousand steps (and thus FLOPs) are close to the other models.

\begin{table}[!h]
\caption{The leading terms of FLOPs in computing attention}
\label{tab:FLOPs}
\centering
\begin{tabular}{l|l}
\hline
\hline
\textbf{Models} & \textbf{FLOPs}  \\
\hline
Standard    & ~ $2n^2 p$  \\
\hline
Big Bird   & ~ $5ndp$ \\
\hline
Performer & ~ $3ndp$ \\
\hline
Nystromformer & ~ $4ndp$ \\
\hline
Linformer & ~ $4ndp$ \\
\hline
Informer & ~ $3ndp$ \\
\hline
Skeinformer  & ~ $4ndp$ \\
\hline
\hline
\end{tabular}
\end{table}

\subsection{Validation loss}
\begin{figure*}[htbp]
\centering
\includegraphics[width=1.0\linewidth]{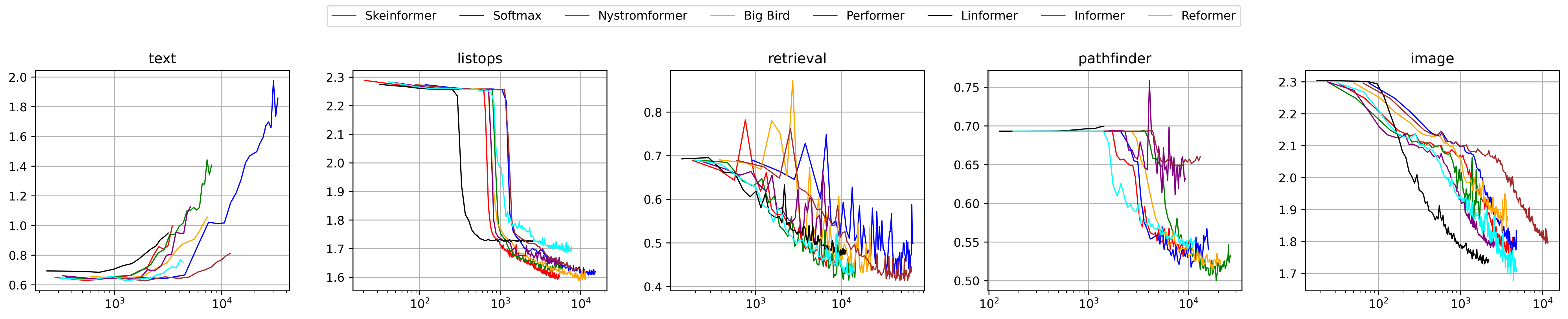}
\caption{Validation loss changes with regard to training time.}
\label{fig:valloss_all}
\end{figure*}
We present the loss decay plots on all tasks in Figure~\ref{fig:valloss_all}.
In the first subplot for the text classification task, we note all the methods quickly overfit the dataset.
In all the other plots, our methods show the ability to both efficiently converge to the long-time limit and find better local minima with lower validation loss.

\section{Proof of Lemma~\ref{lem:pilot}}
\label{sec:lem_pilot}

\begin{proof}
For each column $\mtx{B}^{(i)}$, we first define a discrete random variable $X_i$, 
that with probability $\frac1n$, $X_i = b_{ji}, \forall j \in [n]$, where $b_{ji}$ is the $j$-th element in $\mtx{B}^{(i)}$.
Since all the elements in $\mtx{B}$ are bounded (within the range $[0, 1]$) due to the row normalization in softmax,
we infer that for any $i \in [n], X_i^2 \in [0, 1]$ is a sub-Gaussian random variable with parameter $\sigma = \frac12$ \citep{wainwright2019high}.
Combine the conclusion with the assumption that $\Expect X_i^2 \leq C$, we have 
\begin{align}
\label{eqn:X_i}
\frac{X_i^2}{\Expect X_i^2} \sim \text{sub-Gaussian}\left(\sigma^2=\frac{1}{4 C^2}\right).
\end{align}

Then we uniformly sample $d$ indexes $\{j_k\}_{k=1}^d$'s with replacement, 
and we estimate the squared norm of each column with the unbiased estimator $Y_i = \frac{n}{d} \sum_{k=1}^d b_{j_k i}^2$.
We remark $Y_i$ has the same distribution as $\frac{n}{d} \sum_{k=1}^d X_{i, (k)}^2$, 
where $X_{i, (k)}$'s are $d$ i.i.d. copies of $X_i$.
Therefore through a linear transform of Equation~(\ref{eqn:X_i}) we can derive that 
\begin{align}
\label{eqn:Y_i}
\frac{Y_i}{n \Expect X_i^2} \sim \text{sub-Gaussian}\left(\sigma^2=\frac{1}{4 C^2 d}\right).
\end{align}

Notice different $Y_i$'s may not be independent since they all rely on the same $d$ rows in $\mtx{B}$.
However, we can still apply the maximal sub-Gaussian inequality \citep{boucheron2013concentration} to have:
\begin{align}
\label{eqn:hpbound}
\Prob{\max_{i \in [n]} \abs{\frac{Y_i}{n \Expect X_i^2} - 1} > \frac12} \leq 2n e^{-\frac{C^2 d}{2}}.
\end{align}
If the high probability bound holds that $\max_{i \in [n]} \abs{\frac{Y_i}{n \Expect X_i^2} - 1} \leq \frac12$,
we directly have that our estimators $Y_i \in [\frac12 \|\mtx{B}^{(i)}\|^2, \frac32 \|\mtx{B}^{(i)}\|^2], \forall i \in [n]$.
In that case, the estimated sub-sampling probabilities satisfy that
\begin{align*}
\hat p_i &= \frac{Y_i^{\frac12} \|V_{(i)}\|}
{\sum_{i'=1}^n Y_{i'}^{\frac12} \|V_{(i')}\|}
\geq \frac{\sqrt{1/2} \|\mtx{B}^{(i)}\| \|V_{(i)}\|}
{\sqrt{3/2} \|\mtx{B}^{(i)}\| \|V_{(i')}\|} \\
&= \sqrt{\frac13} p_i, \forall i \in [n],
\end{align*}
where $p_i$'s are the optimal probabilities defined in Equation~(\ref{eqn:beta_prob}) in the main paper.

In that case, to prove the lemma it suffices to pick a big enough sub-sample size $d$ 
such that the right-hand side of Inequality~(\ref{eqn:hpbound}) is smaller than $\delta$.
Simply solving the inequality leads to the desired result $d \geq \frac{2}{C^2} \log(\frac{2n}{\delta})$.
\end{proof}

\end{document}